\documentclass[letterpaper]{article}
\usepackage{aaai19}
\usepackage{times}
\usepackage{helvet}
\usepackage{courier}
\usepackage{url}
\usepackage{graphicx}
\usepackage{subcaption}
\usepackage{booktabs}
\usepackage{multirow}
\usepackage{array}
\usepackage{pgfplots}
\usepackage{amsfonts}
\usepackage{amsmath}
\usepackage{mathtools}
\usepackage{algorithm2e}
\usepackage{tikz}
\usepackage{calc}
\usepackage{textcomp}
\usepackage[para]{threeparttable}
\DeclareMathSizes{10}{8.8}{6.5}{6.5}
\usetikzlibrary{shapes,arrows}
\usetikzlibrary{calc}
\usetikzlibrary{positioning}
\RestyleAlgo{boxruled}
\DeclareMathOperator{\kld}{D_{KL}}
\DeclarePairedDelimiterX{\infdivx}[2]{(}{)}{%
	#1\;\delimsize\|\;#2%
}
\newcommand{\infdiv}{\kld\infdivx}

\DeclareMathOperator*{\argmax}{arg\,max}
\DeclareMathOperator*{\argmin}{arg\,min}

\DeclareRobustCommand\sampleline[1]{
	\tikz\draw[#1] (0,0) (0,\the\dimexpr\fontdimen22\textfont2\relax)
	-- (1em,\the\dimexpr\fontdimen22\textfont2\relax);
}
\pgfplotsset{
	compat=1.3,
	box plot/.style={
		/pgfplots/.cd,
		black,
		only marks,
		mark=-,
		mark size=0.5em,
		/pgfplots/error bars/.cd,
		y dir=plus,
		y explicit,
	},
	box plot box/.style={
		/pgfplots/error bars/draw error bar/.code 2 args={%
			\draw  ##1 -- ++(0.5em,0pt) |- ##2 -- ++(-0.5em,0pt) |- ##1 -- cycle;
		},
		/pgfplots/table/.cd,
		y index=2,
		y error expr={\thisrowno{3}-\thisrowno{2}},
		/pgfplots/box plot
	},
	box plot top whisker/.style={
		/pgfplots/error bars/draw error bar/.code 2 args={%
			\pgfkeysgetvalue{/pgfplots/error bars/error mark}%
			{\pgfplotserrorbarsmark}%
			\pgfkeysgetvalue{/pgfplots/error bars/error mark options}%
			{\pgfplotserrorbarsmarkopts}%
			\path ##1 -- ##2;
		},
		/pgfplots/table/.cd,
		y index=4,
		y error expr={\thisrowno{2}-\thisrowno{4}},
		/pgfplots/box plot
	},
	box plot bottom whisker/.style={
		/pgfplots/error bars/draw error bar/.code 2 args={%
			\pgfkeysgetvalue{/pgfplots/error bars/error mark}%
			{\pgfplotserrorbarsmark}%
			\pgfkeysgetvalue{/pgfplots/error bars/error mark options}%
			{\pgfplotserrorbarsmarkopts}%
			\path ##1 -- ##2;
		},
		/pgfplots/table/.cd,
		y index=5,
		y error expr={\thisrowno{3}-\thisrowno{5}},
		/pgfplots/box plot
	},
	box plot median/.style={
		/pgfplots/box plot
	}
}
\begin{document}
\title{Data Augmentation for Spoken Language Understanding via \\Joint Variational Generation}
\author{Kang Min Yoo, Youhyun Shin, Sang-goo Lee\\
Seoul National University, Seoul 08826, Korea \\
\{kangminyoo, shinu89, sglee\}@europa.snu.ac.kr
}

\nocopyright
\maketitle
\begin{abstract}
	
Data scarcity is one of the main obstacles of domain adaptation in spoken language understanding (SLU) due to the high cost of creating manually tagged SLU datasets. Recent works in neural text generative models, particularly latent variable models such as variational autoencoder (VAE), have shown promising results in regards to generating plausible and natural sentences. In this paper, we propose a novel generative architecture which leverages the generative power of latent variable models to jointly synthesize fully annotated utterances. Our experiments show that existing SLU models trained on the additional synthetic examples achieve performance gains. Our approach not only helps alleviate the data scarcity issue in the SLU task for many datasets but also indiscriminately improves language understanding performances for various SLU models, supported by extensive experiments and rigorous statistical testing.

\end{abstract}

\section{Introduction}

% 1) Data scarcity is a serious problem in building language understanding system
% 2) We have seen significant advancements and growing interest in generative models for nlp, particularly due to the rise of latent variable models such as VAE and GAN
% 3) We propose a novel end-to-end generative model for synthesizing sentences and their respective SLU labels (slot labels and intents), which could improve existing langauge understanding methods and systems by training them with augmented datasets.
% 4) In this paper, we demonstrate superiority of our novel data augmentation approach compared to previous state-of-the-art in data-scarce and general language understanding scenarios

Spoken language understanding (SLU) in current literature refers to the study of models that parse spoken queries into semantic frames. Semantic frames contain pieces of semantic units that best represent the speaker's intentions and are essential for the development of human-machine interfaces, such as virtual assistants.

Scarcity of linguistic resources has been a recurring issue in many NLP tasks such as representation learning \cite{al2013polyglot}, neural machine translation (NMT) \cite{zoph2016transfer}, and SLU \cite{kurata2016labeled}. The issue is especially true for SLU, because creating manually annotated SLU datasets is costly but the domain space that might require new labeled datasets is near infinite.

Even for domains with existing datasets, they might suffer from the data sparsity issue, which have long been plaguing many NLP tasks that require annotated linguistic datasets \cite{lai2015recurrent}. For example, most SLU datasets are not large enough cover all possible data label pairs. Furthermore, biased data collection methods could exacerbate the issue \cite{torralba2011unbiased}.

%such as dropout \cite{srivastava2014dropout} and batch normalization \cite{ioffe2015batch}, which have significantly shaped the way we train deep NLP networks. Transfer learning and multi-task learning have also been popular approaches for mixing out-of-domain knowledge into knowledge-poor models to increase their generalization powers \cite{zoph2016transfer,wu2016google}., 

Recent years, there have been significant advances in variational autoencoders (VAE) \cite{kingma2013auto} and other latent variable models for textual generation \cite{serban2017hierarchical,yu2017seqgan,hu2017toward,li2017adversarial}, prompting investigations into the possibility of improving model performances through generative data augmentation \cite{kafle2017data,kurata2016labeled,hou2018sequence}.

In order to alleviate the data issues, data augmentation (DA) techniques that simply perform class-preserving transformation on data samples have been extensively used extensively \cite{simard2003neural,Krizhevsky2012,fadaee2017data}. However, such DA methods require full supervision and generated datasets lack variety and robustness. To reduce reliance on handcrafted transformation functions, there has been growing interest in leveraging the generative power of latent variable models to facilitate DA. These line of works deserve a category of its own, to which we refer as \textit{generative data augmentation} (GDA). Recent works have explored the idea for the SLU task \cite{kurata2016labeled,hou2018sequence}.

In this paper, we formalize the notion of GDA by developing a general framework for the class of DA techniques in the SLU domain. Upon the framework, we propose a generative model specialized in the generation of SLU datasets. Finally, we wish to demonstrate the effectiveness of our approach through various experiments. In essence, our main contributions are three folds:

\begin{enumerate}
	
	\item \textbf{The Generative DA Framework}: 
	We develop a general framework of generative data augmentation specifically for the SLU task. During formulation, we posit the importance of prior approximation in generation sampling and propose a Monte Carlo-based method. Experiments show that the Monte Carlo-based estimation is superior compared to other approximation methods.

	\item \textbf{A Novel Model for Labeled Language Generation}: 
	We propose a novel generative model for jointly synthesizing spoken utterances and their semantic annotations (slot labels and intents). We show that the synthetic samples generated from the model are not only natural and accurately annotated, but they improve SLU performances by a significant margin when used in the generative data augmentation framework. We also show that our model is better than the previous work \cite{kurata2016labeled}. 

	%Alternatively
	%\item \textbf{A Monte Carlo Method for Latent Variable Sampling}:
	\item \textbf{Substantiation with Extensive Experimentation}: 
	We substantiate the general benefits of generative data augmentation with experiments and statistical testing on various SLU models and datasets. Results show that our approach produces extremely competitive performances for existing SLU models in the ATIS dataset. Our ablation studies also bring some important insights such as the optimal synthetic dataset size to light.
	
\end{enumerate}

\section{Related Work}

\subsubsection{Deep Generative Models and Text Generation}

% This should be moved to introduction ?
Variational autoencoders (VAE) \cite{kingma2013auto,rezende2014stochastic} are deep latent Gaussian models applied with stochastic variational inference, a method which makes the models scalable to large datasets. Since its inception, many variations of the VAE model have been explored in the language domain. Notably, variational recurrent auto-encoders (VRAE)
%, in which both encoders and decoders are based on recurrent neural networks (RNN),
were first proposed by \cite{fabius2014variational}. Generative adversarial networks (GAN) are another class of latent variable models with implicit latent distribution \cite{goodfellow2014generative}. Advances have been made in applying the GAN model to text generation \cite{yu2017seqgan,fedus2018maskgan}. Recently, much attention has been drawn to the tasks of controllable generation and style transfer, which have been successfully explored in \cite{hu2017toward,shen2017style} using variational models.

\subsubsection{Data Augmentation and Regularization}

For data-hungry models, appropriate regularization is necessary to achieve high performance for many tasks. Model regularizers such as dropout \cite{srivastava2014dropout} and batch normalization \cite{ioffe2015batch} are widely accepted techniques to prevent model overfitting and promote noise robustness. Transfer learning is another regularization technique to enhance the generalization power of models that has achieved success across numerous domains and tasks \cite{pan2010survey}.

Data augmentation (DA) is a separate class of regularization methods that create artificial training data to obtain better resulting models. Most DA techniques proposed in the literature can be categorized into either \textit{transformative} or \textit{generative} methods. Transformative data augmentation relies on unparameterized data-transforming functions embued with external knowledge to synthesize new class-preserving data points \cite{dao2018kernel}. Transformative DA is widely used in the vision domain. For example, images are randomly perturbed with linear transformations (rotation, shifting etc.) to boost performances in many vision-related tasks \cite{simard2003neural,Krizhevsky2012}. 
%As a more complex example, text images synthesized from a rendering engine has been used to train a text recognition model intended for real world scenes \cite{jaderberg14synthetic}. Rare word substitution and vocabulary expansion techniques have been explored in the neural machine translation (NMT) and SLU tasks of the language domain \cite{fadaee2017data,khassanov2018unsupervised}.

On the other hand, Generative DA (GDA) exploits the generative power of latent variable models to artificially create convincing data samples. With advances in powerful generative models such as VAEs and GANs, the potential to leverage them for data augmentation has gained much attention recently. Particularly, performance gains from generated datasets have been studied and documented in the VQA task \cite{kafle2017data}, general image classification \cite{ratner2017learning}, and selected SLU tasks \cite{kurata2016labeled,hou2018sequence}. However, relevant researches are hurdled by the architectural and experimental complexities. Nevertheless, our work is the first to explore idea of using variational generative models for DA.

\subsubsection{Spoken Language Understanding}

The SLU task is one of more mature research areas in NLP. Many works have focused on exploring neural architectures for the SLU task. Plain RNNs and LSTMs were first explored in \cite{Mesnil2015,yao2014spoken}. \cite{kurata2016leveraging} proposed sequence-to-sequence (Seq2Seq) models. Hybrid models between RNNs and CRFs were explored in \cite{huang2015bidirectional}. Joint language understanding models that jointly predict slot labels and intents gained significant traction since they had been first proposed in \cite{guo2014joint,goo2018slot}. Some works focused on translating advances in other NLP areas to the SLU task \cite{liu2016attention}.

\tikzset{
	prob/.style    = {draw, thick, rounded corners, rectangle, minimum height=0.8cm, minimum width=1.2cm, node distance=1.2cm, font=\small},
	dataset/.style  = {draw, circle, node distance=1.2cm, minimum size=0.8cm, font=\small},
	input/.style    = {coordinate}, % Input
	output/.style   = {coordinate} % Output
}

\begin{figure}[t]
	\centering
	\begin{tikzpicture}[auto, thick, node distance=2cm, >=triangle 45]
	\draw node [dataset] (original-words) {$\mathbf{w}$};
	\draw node [dataset, below of=original-words] (original-labels) {$\mathbf{s}$};
	\draw node [dataset, below of=original-labels] (original-intents) {$y$};
	\draw [rounded corners, solid, name=original-dataset] ($(original-words.north west) + (-0.25cm, 0.25cm)$) rectangle ($(original-intents.south east) + (0.25cm, -0.25cm)$);
	\path ($(original-words.north west) + (0, 0.2cm)$) -- ($(original-words.north east) + (0, 0.2cm)$) node [midway, above] {$\mathcal{D}$};
	
	\draw node [prob, left=0.7cm of original-words, minimum width=1.2cm] (true-prob) {$p \left( \mathbf{x} \right)$};
	\draw node [prob, left=0.7cm of original-intents, minimum width=1.2cm] (false-prob) {$p^\star \left( \mathbf{x} \right)$};
	%\draw [-latex, dashed, above, font=\footnotesize, shorten >= 0.11cm] (true-prob) -- node {} (original-words);
	\draw [-latex, dashed, above, font=\footnotesize, shorten >= 0.11cm] (false-prob) -- node {} (original-intents);
	\draw [-latex, dashdotted, left, font=\footnotesize] (true-prob) -- node {$\omega_b$} (false-prob);
	
	\draw node [prob, right=0.7cm of original-intents] (data-model) {$\hat{p} \left( \mathbf{x} \right)$};
	\draw node [prob, right=0.7cm of original-words] (better-model) {$\hat{p}^\star \left( \mathbf{x} \right)$};
	\draw [-latex, shorten <= 0.11cm] (original-intents) -- node {} (data-model);
	\draw [-latex, dashdotted, left, font=\footnotesize] (data-model) -- node {$\omega_d$} (better-model);
	
	\draw node [dataset, right=0.7cm of better-model] (new-words) {$\hat{\mathbf{w}}$};
	\draw node [dataset, below of=new-words] (new-labels) {$\hat{\mathbf{s}}$};
	\draw node [dataset, right=0.7cm of data-model] (new-intents) {$\hat{y}$};
	\draw [rounded corners, name=new-dataset] ($(new-words.north west) + (-0.25cm, 0.25cm)$) rectangle ($(new-intents.south east) + (0.25cm, -0.25cm)$);
	\path ($(new-words.north west) + (0, 0.2cm)$) -- ($(new-words.north east) + (0, 0.2cm)$) node [midway, above] {$\hat{\mathcal{D}}$};
	
	\draw [rounded corners, loosely dotted, name=jluva] ($(better-model.north west) + (-0.25cm, 0.6cm)$) rectangle ($(new-intents.south east) + (0.4cm, -0.4cm)$);
	%\path ($(better-model.north west) + (0, 0.11cm)$) -- ($(better-model.north east) + (0, 0.11cm)$) node [midway, above] {GDA};
	
	\draw [-latex, dashed, left, font=\footnotesize, shorten >= 0.11cm] (better-model) -- node {} (new-words);
	
	\draw node [dataset, right=0.7cm of new-words] (aug-words) {$\mathbf{w}'$};
	\draw node [dataset, right=0.7cm of new-labels] (aug-labels) {$\mathbf{s}'$};
	\draw node [dataset, right=0.7cm of new-intents] (aug-intents) {$y'$};
	\draw [rounded corners, name=aug-dataset] ($(aug-words.north west) + (-0.25cm, 0.25cm)$) rectangle ($(aug-intents.south east) + (0.25cm, -0.25cm)$);
	\path ($(aug-words.north west) + (0, 0.2cm)$) -- ($(aug-words.north east) + (0, 0.2cm)$) node [midway, above] {$\mathcal{D}'$};
	
	\draw [-latex, dotted, shorten >= 0.3cm, shorten <= 0.3cm] (original-intents.south east) to [out=335, in=205] (aug-intents.south west);
	\draw [-latex, dotted, shorten >= 0.14cm, shorten <= 0.14cm] (new-labels.east) to [] (aug-labels.west);
	\end{tikzpicture}
	
	\par
	\caption{The general framework for generative language understanding data augmentation. Solid arrows (\sampleline{}) denote training, dashed arrows (\sampleline{dashed}) denote generation, dot-dashed arrows (\sampleline{dashdotted}) denote distortion, and dotted arrows (\sampleline{dotted}) denote data duplication. $\mathcal{D}'$ is the final augmented dataset for training SLU models. The goal of GDA (enclosed in loosely dotted lines) is to recover the true data distribution $p$ through  sampling, as if the samples are drawn from the corrected model distribution.}
	\label{fig:da}
\end{figure}

\section{Model Description}

In this section, we describe our generative data augmentation model and the underlying framework in detail. 

\subsection{Framework Formulation}

We begin with some notations, then we formulate the overall generative data augmentation framework for the spoken language understanding task. 

\subsubsection{Notations}

An utterance $\mathbf{w}$ is a sequence of words $\left( w_1, \ldots, w_{T_i} \right)$, where $T$ is the length of the utterance. For each utterance in a labeled dataset, an equally-long semantic slot sequence $\mathbf{s} = \left( s_1, \ldots, s_T \right)$ exists such that $s_i$ annotates the corresponding word $w_i$. The intent class of the utterance is denoted by $y$. A fully labeled language understanding dataset $\mathcal{D}$ is a collection of utterances and their respective annotations $\left\{ \left( \mathbf{w}_1, \mathbf{s}_1, y_1 \right), \ldots, \left( \mathbf{w}_n, \mathbf{s}_n, y_n \right)\right\}$, where $n$ is the size of the dataset. A data sample in $\mathcal{D}$ is denoted by $\mathbf{x} = \left( \mathbf{w}, \mathbf{s}, y \right)$. The set of all utterances present in $\mathcal{D}$ is denoted by $\mathcal{D}_w = \left\{\mathbf{w}_1, \ldots, \mathbf{w}_n\right\}$. Similarly, the set of slot label sequences and intent classes are denoted by $\mathcal{D}_s$ and $\mathcal{D}_y$.

\subsubsection{Spoken Language Understanding}

A spoken language understanding model is a discriminative model $\mathbf{S}$ fitted on labeled SLU datasets. Specifically, let $\psi$ to denote parameters of the prediction model. Given a training sample $\left(\mathbf{w}, \mathbf{s}, y\right)$, the training objective is as follows:

\begin{equation}
\mathcal{L}_{LU} \left( \psi ; \mathbf{w}, \mathbf{s}, y \right) = -\log p_{\psi} \left( \mathbf{s}, y \middle| \mathbf{w} \right).
\label{eq:lu-nll}
\end{equation}

Given an utterance $\mathbf{w}$, predictions are made by finding the slot label sequence $\hat{\mathbf{s}}$ and the intent class $\hat{y}$ that maximize the  loglikelihood: $\left( \hat{\mathbf{s}}, \hat{y} \right) = \argmax_{\mathbf{s}, y} {\log p_{\psi} \left( \mathbf{s}, y \middle| \mathbf{w} \right)}$.
For non-joint SLU models, $p_\phi$ is factorizable: $p_\phi \left(\mathbf{s}, y \middle| \mathbf{w} \right) = p_\phi \left(\mathbf{s} \middle| \mathbf{w} \right) p_\phi \left( y \middle| \mathbf{w} \right)$. In recent years, joint language understanding has become a popular approach, as studies show a synergetic effect of jointly training slot filling and intent identification \cite{guo2014joint,chen2016syntax}.

%Data augmentation is an effective regularizing technique. 
\subsubsection{Generative Data Augmentation} 

A general framework of generative data augmentation (GDA) is depicted in Figure \ref{fig:da}. Suppose that IID samples $\mathbf{x} \in \mathcal{D}$ were intended to be sampled from a true but unknown language distribution $p \left( \mathbf{x} \right) \in \mathcal{P}$, where $\mathcal{P}$ is the probability function space for $\mathbf{x}$. However, in real world cases, the actual distribution represented by the $\mathcal{D}_w$ could be distorted due to biases introduced during erroneous data collection process or due to under-sampling variance \cite{torralba2011unbiased}. Let such distortion be a function $\omega_b \in \Omega: \mathcal{P} \rightarrow \mathcal{P}$. The distorted data distribution $p^\star = \omega_b \left( p \right)$ diverges from the true distribution $p$, i.e. $d \left( p^\star, p \right) > 0$ where $d$ is some statistical distance measure such as KL-divergence. 

An ideal GDA counteracts the bias-introducing function $\omega_b$ and unearths the true distribution $p$ through unsupervised explorative sampling. Suppose that a joint language understanding model $\hat{p} \left( \mathbf{x} \right)$ is trained on $\mathbf{x} \sim p^\star \left( \mathbf{x} \right)$. Without the loss of generality, suppose that the model is expressive enough to perfectly capture the underlying distribution, i.e. $\hat{p} = p^\star$. We collect $m$ samples $\hat{\mathcal{D}} = \left\{\hat{\mathbf{x}}_1,\ldots,\hat{\mathbf{x}}_m\right\}$ drawn from $\hat{p} \left( \mathbf{x} \right)$ and combine them with the original dataset $\mathcal{D}$ to form an augmented dataset $\mathcal{D}'$ of size $n + m$. Na\"ive DA will not yield better SLU results as synthetic data samples $\hat{\mathbf{x}}$ follow the distorted data distribution $p^\star$ in the best case. However, an ideal explorative sampling method could distort the sampling distribution, as if $\hat{\mathbf{x}}$ were sampled from another distribution $\hat{p}^\star$, such that the new distribution is closer to the true distribution (i.e. $d \left( \hat{p}^\star, p \right) < d \left(\hat{p}, p\right)$). There exists a distortion function $\omega_d$ such that $\hat{p}^\star = \omega_d \left( \hat{p} \right)$. The ideal sampling method can be seen as a corrective function $\omega_d$ that undos the effect of $\omega_b$. In this paper, we propose and investigate different sampling methods $\omega_d$ for the maximal DA effect. These methods are described in model description sections. The implementation details are covered in the experiments sections.%Due to the regularizing power of auto-encoding latent variable models, the new dataset $\mathcal{D}'$ generated from our joint generative model is theorized to represent a better data distribution. The experimental results (presented in subsequent sections) support our hypothesis.

\begin{algorithm}[!ht]
	\SetKwData{Left}{left}\SetKwData{This}{this}\SetKwData{Up}{up}
	\SetKwFunction{Union}{Union}\SetKwFunction{FindCompress}{FindCompress}
	\SetKwInOut{Input}{input}\SetKwInOut{Output}{output}
	\SetKwInOut{Given}{given}
	\Input{a sufficiently large number $m$}
	\Given{$\mathcal{D}_w$, $\theta$, $\phi$}
	\Output{synthetic utterance list $\mathbf{U}$}
	initialize $\mathbf{U}$ as an empty list\;
	\While{$\mathbf{U}$ has less than $m$ samples}{
		sample a real utterance $\mathbf{w}$ from  $\mathcal{D}_w$\;
		estimate the mean $\bar{\mathbf{z}}$ of the posterior $q_\phi \left(\mathbf{z} \middle| \mathbf{w}\right)$\;
		sample $\hat{\mathbf{w}}$ from the likelihood $p_\theta \left( \mathbf{w} \middle| \mathbf{\bar{z}} \right)$\;
		append $\hat{\mathbf{w}}$ to $\mathbf{U}$\;
	}
	\Return{$\mathbf{U}$}
	\caption{Monte Carlo posterior sampling.}
	\label{alg:posterior-sampling}
\end{algorithm}

\subsection{Joint Generative Model}

In this subsection, we describe our generative model in detail. We begin with a standard VAE \cite{kingma2013auto} applied to utterances, then we extend the model by allowing it to generate other labels in a joint fashion.

\subsubsection{Standard VAE}
 
VAEs are latent variable models applied with amortized variational inference. Let $\theta$ be the parameters of the generator network (i.e. the decoder network), and let $\phi$ be the parameters of the recognition network (i.e. the encoder network). Specifically in the case of utterance learning, the goal is to maximize the log likelihood of sample utterances $\mathbf{w}$ in the dataset $\log p \left( \mathbf{w} \right) = \log \int p \left(\mathbf{w},\mathbf{z}\right) d\mathbf{z}$. However, since the marginalization is computationally intractable, we introduce a proxy network $q_\phi \left( \mathbf{z} \middle| \mathbf{w} \right)$ and subsequently minimize a training objective based on evidence lower bound (ELBO): 

\begin{align}
	\begin{split} 
		\mathcal{L}_{VAE} \left( \theta, \phi; \mathbf{w} \right) = 
		&\infdiv{q_\phi \left( \mathbf{z} \middle| \mathbf{w} \right)}{p \left( \mathbf{z} \right)}\\
		&-\mathbb{E}_{\mathbf{z} \sim q_\phi \left( \mathbf{z} \middle| \mathbf{w} \right)} \left[\log p_\theta \left(\mathbf{w} \middle| \mathbf{z}\right)\right]
	\end{split}
	\label{eq:elbo}
\end{align} 

In Equation \ref{eq:elbo}, the proxy distribution $q_\phi$ is kept close to the prior $p\left(\mathbf{z}\right)$, which we assume to be the standard multivariate Gaussian. Since the KL-divergence term is always positive, $\mathcal{L}_{VAE}$ is the upper bound for the reconstruction error under the particular choice of a proxy distribution $q_\phi$. The proposed generative model is based on VRAEs, in which the posterior of a sequence factorizes over sequence elements (i.e. words) based on the Markov Chain assumption: $p_\theta \left( \mathbf{w} \middle| \mathbf{z} \right) = \prod_{i=1}^T p_\theta \left( w_i | w_1,\ldots,w_{i-1},\mathbf{z} \right)$. VAEs can be optimized using gradient-descent methods with the reparameterization trick \cite{kingma2013auto}.

\tikzset{
	vector/.style = {draw, rectangle, minimum height=0.9cm, minimum width=0.15cm, node distance=0.6cm, line width=0.5pt},
	point/.style = {draw, circle, inner sep=0pt, minimum size=0.1cm, line width=0.3pt},
	label/.style = {font=\footnotesize}
}

\begin{figure}[t]
	\centering
	\begin{tikzpicture}[auto]
	\draw node [vector] (h1enc) {};
	\draw node [vector, right of=h1enc] (h2enc) {};
	\draw node [vector, right of=h2enc] (h3enc) {};
	\draw node [vector, right of=h3enc] (h4enc) {};
	\draw node [vector, right of=h4enc] (h5enc) {};
	
	\draw node [point, below=0.35cm of h1enc] (w1in) {};
	\draw node [point, below=0.35cm of h2enc] (w2in) {};
	\draw node [point, below=0.35cm of h3enc] (w3in) {};
	\draw node [point, below=0.35cm of h4enc] (w4in) {};
	\draw node [point, below=0.35cm of h5enc] (w5in) {};
	
	\draw node [label, below=0.05cm of w1in] {$w_1$};
	\draw node [label, below=0.05cm of w2in] {$w_2$};
	\draw node [label, below=0.05cm of w3in] {$w_3$};
	\draw node [label, below=0.05cm of w4in] {$w_4$};
	\draw node [label, below=0.05cm of w5in] {$w_5$};
	
	\draw [-stealth] (w1in) -- node {} (h1enc);
	\draw [-stealth] (w2in) -- node {} (h2enc);
	\draw [-stealth] (w3in) -- node {} (h3enc);
	\draw [-stealth] (w4in) -- node {} (h4enc);
	\draw [-stealth] (w5in) -- node {} (h5enc);
	
	\draw [-stealth] (h1enc.55) -- node {} (h2enc.125);
	\draw [-stealth] (h2enc.55) -- node {} (h3enc.125);
	\draw [-stealth] (h3enc.55) -- node {} (h4enc.125);
	\draw [-stealth] (h4enc.55) -- node {} (h5enc.125);
	\draw [-stealth] (h5enc.235) -- node {} (h4enc.305);
	\draw [-stealth] (h4enc.235) -- node {} (h3enc.305);
	\draw [-stealth] (h3enc.235) -- node {} (h2enc.305);
	\draw [-stealth] (h2enc.235) -- node {} (h1enc.305);
	
	\draw [rounded corners, dashed, name=encoder] ($(h1enc.north west) + (-0.20cm, 0.20cm)$) rectangle ($(h5enc.south east) + (0.20cm, -0.20cm)$);
	\path ($(h1enc.north west) + (-0.15cm, 0.2cm)$) -- ($(h5enc.north east) + (0.1cm, 0.2cm)$) node [near start, above, font=\footnotesize] {Encoder};
	
	\draw node [vector, above right=-0.4cm and 0.5cm of h5enc] (mu) {};
	\draw node [vector, below right=-0.4cm and 0.5cm of h5enc] (std) {};
	\draw node [label, above left=-0.5cm and -0.05cm of mu] () {$\mu$};
	\draw node [label, below left=-0.5cm and -0.05cm of std] () {$\sigma$};
	\draw [-stealth, shorten <= 0.26cm] (h5enc.70) -- node {} (mu);
	\draw [-stealth, shorten <= 0.26cm] (h5enc.290) -- node {} (std);
	
	\draw node [vector, right=1.2cm of h5enc] (z) {};
	\draw node [label, left=-0.06cm of z] {$z$};
	\draw [-stealth, dotted] (mu) -- node {} (z.110);
	\draw [-stealth, dotted] (std) -- node {} (z.250);
	
	\draw node [vector, right of=z] (h1wdec) {};
	\draw node [vector, right of=h1wdec] (h2wdec) {};
	\draw node [vector, right of=h2wdec] (h3wdec) {};
	\draw node [vector, right of=h3wdec] (h4wdec) {};
	\draw node [vector, right of=h4wdec] (h5wdec) {};
	
	\draw [-stealth] (z) -- node {} (h1wdec);
	\draw [-stealth] (h1wdec) -- node {} (h2wdec);
	\draw [-stealth] (h2wdec) -- node {} (h3wdec);
	\draw [-stealth] (h3wdec) -- node {} (h4wdec);
	\draw [-stealth] (h4wdec) -- node {} (h5wdec);
	
	\draw node [point, above=0.35cm of h1wdec] (w1out) {};
	\draw node [point, above=0.35cm of h2wdec] (w2out) {};
	\draw node [point, above=0.35cm of h3wdec] (w3out) {};
	\draw node [point, above=0.35cm of h4wdec] (w4out) {};
	\draw node [point, above=0.35cm of h5wdec] (w5out) {};
	
	\draw node [label, right=-0.1cm of w1out] {$\hat{w}_1$};
	\draw node [label, right=-0.1cm of w2out] {$\hat{w}_2$};
	\draw node [label, right=-0.1cm of w3out] {$\hat{w}_3$};
	\draw node [label, right=-0.1cm of w4out] {$\hat{w}_4$};
	\draw node [label, right=-0.1cm of w5out] {$\hat{w}_5$};
	
	\draw [-stealth] (h1wdec) -- node {} (w1out);
	\draw [-stealth] (h2wdec) -- node {} (w2out);
	\draw [-stealth] (h3wdec) -- node {} (w3out);
	\draw [-stealth] (h4wdec) -- node {} (w4out);
	\draw [-stealth] (h5wdec) -- node {} (w5out);
	\draw [-stealth] (h1wdec.north) |- ++(0, 0.1cm) -| ++(0.3cm, 0) |- ++(0, -1.25cm) -| node {} (h2wdec.south);
	\draw [-stealth] (h2wdec.north) |- ++(0, 0.1cm) -| ++(0.3cm, 0) |- ++(0, -1.25cm) -| node {} (h3wdec.south);
	\draw [-stealth] (h3wdec.north) |- ++(0, 0.1cm) -| ++(0.3cm, 0) |- ++(0, -1.25cm) -| node {} (h4wdec.south);
	\draw [-stealth] (h4wdec.north) |- ++(0, 0.1cm) -| ++(0.3cm, 0) |- ++(0, -1.25cm) -| node {} (h5wdec.south);
	
	\draw node [vector, above=0.8cm of h1wdec] (h1sdec) {};
	\draw node [vector, above=0.8cm of h2wdec] (h2sdec) {};
	\draw node [vector, above=0.8cm of h3wdec] (h3sdec) {};
	\draw node [vector, above=0.8cm of h4wdec] (h4sdec) {};
	\draw node [vector, above=0.8cm of h5wdec] (h5sdec) {};
	
	\draw [-stealth] (w1out) -- node {} (h1sdec.south);
	\draw [-stealth] (w2out) -- node {} (h2sdec.south);
	\draw [-stealth] (w3out) -- node {} (h3sdec.south);
	\draw [-stealth] (w4out) -- node {} (h4sdec.south);
	\draw [-stealth] (w5out) -- node {} (h5sdec.south);
	
	\draw [-stealth] (h1sdec) -- node {} (h2sdec);
	\draw [-stealth] (h2sdec) -- node {} (h3sdec);
	\draw [-stealth] (h3sdec) -- node {} (h4sdec);
	\draw [-stealth] (h4sdec) -- node {} (h5sdec);
	
	\draw [-stealth] (z) |- node {} (h1sdec);
	
	\draw node [point, above=0.35cm of h1sdec] (s1out) {};
	\draw node [point, above=0.35cm of h2sdec] (s2out) {};
	\draw node [point, above=0.35cm of h3sdec] (s3out) {};
	\draw node [point, above=0.35cm of h4sdec] (s4out) {};
	\draw node [point, above=0.35cm of h5sdec] (s5out) {};
	
	\draw node [label, above=0cm of s1out] {$\hat{s}_1$};
	\draw node [label, above=0cm of s2out] {$\hat{s}_2$};
	\draw node [label, above=0cm of s3out] {$\hat{s}_3$};
	\draw node [label, above=0cm of s4out] {$\hat{s}_4$};
	\draw node [label, above=0cm of s5out] {$\hat{s}_5$};
	
	\draw [-stealth] (h1sdec) -- node {} (s1out);
	\draw [-stealth] (h2sdec) -- node {} (s2out);
	\draw [-stealth] (h3sdec) -- node {} (s3out);
	\draw [-stealth] (h4sdec) -- node {} (s4out);
	\draw [-stealth] (h5sdec) -- node {} (s5out);
	
	\draw node [vector, below=0.6cm of h1wdec] (h1idec) {};
	\draw node [vector, below=0.6cm of h2wdec] (h2idec) {};
	\draw node [vector, below=0.6cm of h3wdec] (h3idec) {};
	\draw node [vector, below=0.6cm of h4wdec] (h4idec) {};
	\draw node [vector, below=0.6cm of h5wdec] (h5idec) {};
	
	\draw [-stealth] (h1idec) -- node {} (h2idec);
	\draw [-stealth] (h2idec) -- node {} (h3idec);
	\draw [-stealth] (h3idec) -- node {} (h4idec);
	\draw [-stealth] (h4idec) -- node {} (h5idec);
	
	\draw [-stealth] (z.south) |- node {} (h1idec);
	
	\draw node [point, right=0.3cm of h5idec] (iout) {};
	\draw node [label, above=0.0cm of iout] {$\hat{y}$};
	\draw [-stealth] (h5idec) -- node {} (iout);
	
	\draw [-stealth] (h1wdec.north) |- ++(0, 0.1cm) -| ++(0.3cm, 0) |- ++(0, -1.35cm) -| node {} (h1idec.north);
	\draw [-stealth] (h2wdec.north) |- ++(0, 0.1cm) -| ++(0.3cm, 0) |- ++(0, -1.35cm) -| node {} (h2idec.north);
	\draw [-stealth] (h3wdec.north) |- ++(0, 0.1cm) -| ++(0.3cm, 0) |- ++(0, -1.35cm) -| node {} (h3idec.north);
	\draw [-stealth] (h4wdec.north) |- ++(0, 0.1cm) -| ++(0.3cm, 0) |- ++(0, -1.35cm) -| node {} (h4idec.north);
	\draw [-stealth] (h5wdec.north) |- ++(0, 0.1cm) -| ++(0.3cm, 0) |- ++(0, -1.35cm) -| node {} (h5idec.north);
	
	\draw [rounded corners, dashed, name=encoder] ($(h1sdec.north west) + (-0.22cm, 0.15cm)$) rectangle ($(h5idec.south east) + (0.55cm, -0.2cm)$);
	\path ($(h5sdec.north east) + (0.1cm, 0.3cm)$) -- ($(h5idec.north east) + (0.2cm, 0cm)$) node [near start, above, font=\footnotesize, rotate=270] {Decoders};
	
	\end{tikzpicture}
	\caption{Joint language understanding variational autoencoder (JLUVA). The VAE model consists of a BiLSTM-Max encoder and three uni-directional decoders. Note that the fully connected layers and embedding layers are omitted for clarity.}
	\label{fig:jluva}
\end{figure}
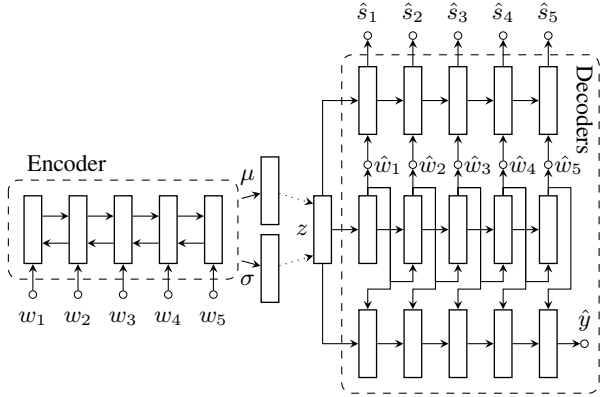
6
\subsubsection{The Sampling Problem}

Given the parameters $\theta_\mathcal{D}$ and $\phi_\mathcal{D}$ that are optimized for all $\mathbf{w} \in \mathcal{D}_w$, our goal is to sample plausible utterances $\hat{\mathbf{w}}$ from the distribution of $\mathbf{w}$ believed by the model:

\begin{equation}
	\hat{\mathbf{w}} \sim p_{\theta_\mathcal{D},\phi_\mathcal{D}}\left(\mathbf{w}\right) =\int p_{\theta_\mathcal{D}}\left(\mathbf{w} \middle| \mathbf{z}\right) p_{\theta_\mathcal{D},\phi_\mathcal{D}}\left(\mathbf{z}\right)d\mathbf{z}
	\label{eq:general-sampling}
\end{equation}

As evident in Equation \ref{eq:general-sampling}, the marginal likelihood estimation requires us to infer the marginal probability of the latent variable  $p_{\theta_\mathcal{D},\phi_\mathcal{D}}\left(\mathbf{z}\right)$, which can be estimated by marginalizing the joint probability from the recognition network.

\begin{equation}
p_{\theta_\mathcal{D},\phi_\mathcal{D}}\left(\mathbf{z}\right) = \mathbb{E}_{\mathbf{w} \sim p \left( \mathbf{w} \right)} \left[q_{\phi_\mathcal{D}}\left(\mathbf{z}\middle|\mathbf{w}\right)\right]
\label{eq:marginalize-encoder}
\end{equation}

However, Equation \ref{eq:marginalize-encoder} cannot be solved analytically, as the true distribution of $w$ is unknown. Hence, some form of approximation is required to sample utterances from the latent variable model. The approximation approach will likely have an impact on the quality of generated utterances, thereby determine the effect of data augmentation. Here, we describe two options. 

The first is to approximate the marginal probability of the latent variable with the prior $p \left( \mathbf{z}  \right)$, the standard multivariate Gaussian. However, this na\"ive approximation will likely yield homogeneous and uninteresting utterances due to over-simplication of the latent variable space. In real world scenarios, the KLD loss term in Equation \ref{eq:elbo} is still large after convergence.

% which indicates the existence of discrepancies between the latent space distribution encoded by the recognition network and the ideal distribution we wish to impose \cite{bowman2016generating}. 

Alternatively, the other option is to approximate using the Monte Carlo method. Under the Monte Carlo approach (Algorithm \ref{alg:posterior-sampling}), the marginal likelihood is calculated deterministically for each utterance $w$ sampled from the dataset $\mathcal{D}$. According to the law of large numbers, the marginal likelihood $p_{\theta_\mathcal{D},\phi_\mathcal{D}} \left( \mathbf{w} \right)$ converges to the empirical mean, thereby providing an unbiased distribution for sampling $\mathbf{w}$.

%In practice, due to the variance in sampling $\mathbf{w}$ from $\mathcal{D}$, Algorithm \ref{alg:posterior-sampling} might require unrealistically large number of trials. Slow convergence to the empirical mean could introduce biases to the augmented dataset, which could deteriorate the performance of SLU models trained on the dataset. To inhibit variance, we enforce a variance-reducing technique during likelihood sampling. Specifically, during likelihood sampling $\hat{\mathbf{w}} \sim p_\theta \left( \mathbf{w} \middle| \mathbf{z} \right)$ we only sample utterances from top-$k_b$ most likely utterances, which are identified by the beam search algorithm. 

\subsubsection{Exploratory Sampling}

In our general framework for GDA, remind that the sampling method is required to be exploratory, such that the biases in datasets are counteracted. translating to better performances in resulting models. Hence, an ideal exploratory sampling approach is unbiased but has increased sampling variance. Intuitively, we can sample the latent variable $\mathbf{z}$ from the Gaussian encoded by the recognizer in place of analytically estimating the mean in Algorithm \ref{alg:posterior-sampling}. Suppose that $\boldsymbol{\mu}$ and $\boldsymbol{\sigma}$ are mean and standard deviation vectors encoded by the recognizer. Then we sample $\mathbf{z}$ from $\mathcal{N} \left(\boldsymbol{\mu}\left(\mathbf{w}\right), \lambda_s \cdot \boldsymbol{\sigma}\left(\mathbf{w}\right)\right)$, where the scaling hyperparameter $\lambda_s$ controls the level of exploration exhibited by the generator. This unbiased empirical estimation of the posterior helps generate realistic but more varied utterances.

\subsubsection{Joint Language Understanding VAE}

Starting from a VAE for encoding and decoding utterances, Joint Language Understanding VAE (JLUVA) extends the model by predicting slot labels and intent classes. The generation of slot labels and intents are conditioned on the latent variable $\mathbf{z}$ and the generated utterance $\hat{\mathbf{w}}$ (Figure \ref{fig:jluva}). The benefits of having conditional dependence on $\mathbf{z}$ during labeling is documented in \cite{kurata2016leveraging}. The modified training objective for the language understanding task is as follows.

\begin{table}
	\centering
	\begin{tabular}{lcccccc}
		\hline
		Dataset			        & \#Splits	& Train		& Val		& Test		\\
		\hline\hline
		ATIS-small	    & 35		& 127 - 128	& 500		& 893		\\
		ATIS-medium	    & 9			& 497 - 498 & 500		& 893		\\ 
		ATIS	                & 1			& 4,478		& 500		& 893		\\
		Snips			        & 1			& 13,084	& 700		& 700		\\
		MIT Movie Eng	    & 1			& 8,798		& 977		& 2,443		\\
		MIT Movie Trivia  & 1			& 7,035		& 781		& 1,953		\\
		MIT Restaurant	    & 1			& 6,894		& 766		& 1,521		\\
		\hline
	\end{tabular}
	\caption{Dataset statistics. Training sets of ATIS (Small) and ATIS (Medium) have been chunked from the training set of ATIS (Full). }
	\label{tab:datastats}
\end{table}

\begin{equation}
\mathcal{L}_{LU} \left( \phi, \psi ; \mathbf{w}, \mathbf{s}, y \right) = -\mathbb{E}_{\mathbf{z} \sim q_\phi} \left[ \log p_{\psi} \left( \mathbf{s}, y \middle| \hat{\mathbf{w}}, \mathbf{z} \right) \right]
\label{eq:lu-new-nll}
\end{equation}

The joint training objective of the entire model is specified in terms of the training objective of the VAE component (Equation \ref{eq:elbo}) and the negative log-likelihood of the discriminatory component (Equation \ref{eq:lu-new-nll}):

\begin{table*} 
	\centering
	\begin{threeparttable}
		\begin{tabular}{l|ccc|ccc|ccc}
			\hline 
			& \multicolumn{3}{c}{Slot Filling (F1)}					& \multicolumn{3}{c}{Intent (F1)} 					& \multicolumn{3}{c}{Semantic (Acc.)}  \\
			\cline{2-10}
			Model + Sampling Approach 														      & Small	 			& Med. 				& Full 				& Small 			& Med. 				                                                & Full 		& Small 			& Med. 			& Full  \\
			\hline \hline
			\begin{tabular}{@{}@{} l}Baseline (No Augmentation) \end{tabular}  				      & 72.57\tnote{\ddag} 	& 88.28\tnote{\ddag}	& 95.34 			& 82.65 			& 90.59\tnote{\dag} 	& 97.21 & 35.09\tnote{\ddag} 	& 65.18\tnote{\ddag}	& 85.95 \\
			\begin{tabular}{@{}@{} l}Encoder-Decoder + Additive\tnote{*} \end{tabular} 	          & 74.79\tnote{\dag} 	& 89.13\tnote{\ddag}	& 95.20& - 			& - 				& - 		& - 				& - 				& - \\ % koala
			\begin{tabular}{@{}@{} l}JLUVA + Additive (Ours) \end{tabular} 					      & 74.14\tnote{\ddag} 	& 89.13\tnote{\ddag}	& 95.40 			& 83.46 			& \textbf{90.97} 	& 97.04 	& 38.58			& 66.75 			& 85.81 \\ % whaleshark
			\begin{tabular}{@{}@{} l}JLUVA + Standard Gaussian (Ours) \end{tabular} 	          & 70.72\tnote{\ddag}	& 86.90\tnote{\ddag}  & 94.91\tnote{\ddag} 	& 78.67\tnote{\ddag}	& 86.90\tnote{\ddag}	& 96.90 	& 32.46\tnote{\ddag}	& 61.12\tnote{\ddag}	& 84.62\tnote{\ddag} \\ % seahorse
			\begin{tabular}{@{}@{} l}JLUVA + Posterior (Ours) \end{tabular} 				      & \textbf{74.92} 	& \textbf{89.27} 	& \textbf{95.51} 			& \textbf{83.65} 	& 90.95 			& \textbf{97.24} 	& \textbf{39.43} 	& \textbf{67.05} 	& \textbf{86.26} \\ % seaelephant
			\hline
		\end{tabular}
		\begin{tablenotes}
			\item[*]{\footnotesize \cite{kurata2016labeled}}
			\item[\dag]{\footnotesize $p<0.1$}
			\item[\ddag]{\footnotesize $p<0.01$}
		\end{tablenotes}
	\end{threeparttable}
	\caption{
		Data scarcity results for the ATIS dataset. We use the baseline BiLSTM model as the control SLU model. Results are averaged over multiple runs and compared to the best of our approaches (JLUVA + Posterior). The differences are tested for statistical significance.
	}
	\label{tab:data-scarce}
\end{table*}

\begin{align}
\begin{split}
\mathcal{L} \left( \theta, \phi, \psi; \mathbf{w}, \mathbf{s}, y \right) = &\infdiv{q_\phi \left( \mathbf{z} \middle| \mathbf{w} \right)}{p_\theta \left( \mathbf{z} \middle| \mathbf{w} \right)}\\
&-\mathbb{E}_{\mathbf{z} \sim q_\phi} \left[\log p_\theta \left(\mathbf{w} \middle| \mathbf{z}\right)\right]  \\
&-\mathbb{E}_{\mathbf{z} \sim q_\phi} \left[ \log p_{\psi} \left( \mathbf{s}, y \middle| \hat{\mathbf{w}}, \mathbf{z} \right) \right]
\end{split}
\label{eq:jluva-loss}
\end{align}

We obtain the optimal parameters $\theta^*, \phi^*, \psi^*$ by minimizing Equation \ref{eq:jluva-loss} (i.e. $\argmin_{\theta, \phi, \psi}\mathcal{L}$) with respect to a real dataset $\mathcal{D}$. During the data generation process, we sample $z^\star$ from an approximated prior $p^\star \left( \mathbf{z} \right)$ which depends on the approximation strategy (e.g. posterior sampling). Then we perform inference on the posterior network $p_\theta \left( \mathbf{w} \middle| \mathbf{z^\star} \right)$ to estimate the language distribution. A synthetic utterance $\hat{\mathbf{w}}$ is sampled from said distribution and is used to infer the slot label and intent distribution from the relevant networks, i.e. $p \left( \mathbf{s}, y \middle| \mathbf{z}, \hat{\mathbf{w}} \right)$. The most probable $\hat{\mathbf{s}}$ and $\hat{y}$ are combined with $\hat{\mathbf{w}}$ to form a generated sample set $\left( \hat{\mathbf{w}}, \hat{\mathbf{s}}, \hat{y} \right)$. This generation process is repeated until sufficient synthetic data samples are collected.

\section{Experiments}
\label{sec:exp}

In this section, we outline the design, execution, results and analysis of all experiments pertaining to testing the effectiveness of our GDA approach.

\subsection{Datasets}

In this paper, we carry out experiments on the following language understanding datasets. 

\begin{itemize}
	
	\item \textbf{ATIS}: Airline Travel Information System (ATIS) \cite{hemphill1990atis} is a representative dataset in the SLU task, providing well-founded comparative environment for our experiments. 
	
	\item \textbf{Snips}: The snips dataset is an open source virtual-assistant corpus. The dataset contains user queries from various domains such as manipulating playlists or booking restaurants.
	
	\item \textbf{MIT Restaurant (MR)}: This single-domain dataset specializes in spoken queries related to booking restaurants. 
	
	\item \textbf{MIT Movie}: The MIT movie corpus consists of two single-domain datasets: the movie eng (ME) and movie trivia (MT) datasets. While both datasets contain queries about film information, 
	%queries in the movie eng dataset are simpler,
	% (e.g. "who directed indiana jones?")
	the trivia queries are more complex and specific. 
	%(e.g. "what is this 1968 horror movie that sparked the zombie craze?"). 

\end{itemize}

All of the datasets are annotated with slot labels and intent classes except the MIT datasets. The detailed statistics of each dataset are shown in Table \ref{tab:datastats}. In order to simulate a data scarce environment (similar to the experimental design proposed in \cite{chen2016syntax}), we randomly chunk the ATIS training set into equal-sized smaller splits. For the small dataset the training set is chunked into 35 pieces, and for the medium dataset it is chunked into 9 pieces. The sizes of the small and medium training splits approximately correspond those mentioned in the previous work \cite{chen2016syntax}.

%\subsection{Baselines}

%In this subsection, we present the baselines for our studies. The baseline for generative models a deterministic encoder-decoder for slot labels, 
%Since our approach consists of several smaller components (i.e. the generative model, the sampling approach, the SLU model), elaboration on the baseline settings for all parts of the experiments.

%\subsubsection{Generative Models}

%As one of the baselines for our generative model, 

%\subsubsection{Sampling Approaches}

%We compare different approximation approaches for the marginal latent variable distribution $p_{\theta,\phi} \left( \mathbf{z} \right)$.

\subsection{Experimental Settings}

Here, we describe the methodological and implementation details for testing the GDA approach under the framework.

\subsubsection{General Experimental Flow}

Since we observe a high variance in performance gains among different runs of the same generative model (e.g. Figure \ref{fig:ablation}), we need to approach the experimental designs with a more conservative stance. The general experimental methodology is as follows.

\begin{itemize}
	\item $N_G$ identical generative models are trained with different initial seeds on the same training split.
	\item $m$ utterance samples are drawn from each model to create $N_G$ augmented datasets $\mathcal{D}'_1,\ldots,\mathcal{D}_{N_G}'$.
	\item $N_L$ identical SLU models are train for \textit{each} augmented dataset $\mathcal{D}'$. All models are validated against the evaluation results on the same validation split. Best model from each SLU model is evaluated on the test set.
	\item We collect the statistics of all $N_G \times N_L$ results and perform comparative analyses.
\end{itemize}

\subsubsection{Implementation Details}

For all of our models, the word embeddings $W_w$, slot label embeddings $W_s$, and intent embeddings $W_y$ dimensions were 300, 200, and 100 respectively and they were trained with the rest of the network. The word embeddings had been initialized with the GloVe vectors \cite{pennington2014glove}. 

For the generative model, the encoder is a single-layer BiLSTM-Max model \cite{conneau2017supervised}, which encodes the word embeddings of word tokens $w_i \in \mathbf{w}$ in both directions and produces the final hidden state by applying max-pooling-over-time on combined encoder hidden outputs $\mathbf{h}^{\left(e\right)}_1,\ldots,\mathbf{h}^{\left(e\right)}_T$ (1024 hidden dimensions). The decoders are uni-directional single-layer LSTMs with the same hidden dimensions (1024). Let $\mathbf{h}^{\left(w\right)}_t$, $\mathbf{h}^{\left(s\right)}_t$, and $\mathbf{h}^{\left(y\right)}_t$ be the hidden outputs of word, slot label, and intent decoders at time step $t$ respectively. We perform dot products between respective embeddings and the hidden outputs to obtain logits (e.g. $\mathbf{o}^{\left(w\right)}_t = W_w \mathbf{h}^{\left(w\right)}_t$ etc.). The likelihood of each token at each time step $t$ is obtained by applying the softmax on the logits:

$$p \left( w_t \middle| \mathbf{w}_{<t}, \mathbf{z} \right) = \frac{e^{o^{\left(w\right)}_{t,w_t}}}{\sum_{w' \in V_w}{ e^{o^{\left(w\right)}_{t,w'}}}}.$$

Where $V_w$ is the vocabulary set of utterance words. During generation, the beam search algorithm is used to search for the most likely sequence candidates using the conditional token distributions. The beam search size was set to 15 and the utterances were sampled from top-1 ($k_b$) candidate(s) to reduce variance. Exploratory hyperparameter $\lambda_s$ was 0.18. 

To feasibly train the model, we employ the teacher-forcing strategy, in which the LU network is trained on the ground truth utterance $\mathbf{w}$ instead of the predicted sequence $\hat{\mathbf{w}}$. We applied KLD annealing and the decoder word dropout \cite{bowman2016generating}.
%to attain healthy training of the variational model. 
KLD annealing rate ($k_d$) was 0.03 and word dropout rate $p_w$ was 0.5. We used Adam \cite{kingma2014adam} optimizer with 0.001 initial learning rate. 

\subsubsection{SLU Models}

\tikzset{
	vector/.style = {draw, rectangle, minimum height=1.2cm, minimum width=0.15cm, node distance=0.6cm, line width=0.5pt},
	point/.style = {draw, circle, inner sep=0pt, minimum size=0.1cm, line width=0.3pt},
	label/.style = {font=\footnotesize}
}

For the baseline SLU model, we implemented a simple BiLSTM model. A bidirectional LSTM cell encodes an utterance into a fixed size representation  $h$. A fully connected layer translates the hidden outputs $h_t$ of the BiLSTM to slot scores for all time step $t$. The softmax function is applied to the logits to produce $p \left( s_t \middle| \mathbf{w}_{\leq t} \right)$. The final hidden representation $h$ of the input utterance is obtained by applying max-pooling-over-time on all hidden outputs. Another fully connected layer and a softmax function maps $h_t$ to the intent distribution $p \left( y \middle| \mathbf{w} \right)$. This simple baseline was able to achieve 95.32 in the slot filling f1-score.

\begin{table}
	\centering
	\begin{threeparttable}
		\begin{tabular}{l|ccc|ccc}
			\hline 
			& \multicolumn{3}{c}{Slot-Gated (Full)} 					    & \multicolumn{3}{c}{Slot-Gated (Intent)} \\
			\cline{2-7}
			Dataset			& Slot			        & Intent 		        & SF		            & Slot			        & Intent     		& SF		                    \\
			\hline \hline                                                                                                   
			ATIS 			& 95.3\tnote{\ddag}	& 94.9\tnote{\ddag}	& 84.3\tnote{\ddag}	& 95.4\tnote{\ddag}	& 94.7\tnote{\ddag}& 83.5\tnote{\ddag}			\\ %Max Full: 95.66
			ATIS+ 			& \textbf{95.7}	    & \textbf{95.6}        & \textbf{85.4}		& \textbf{95.6}		& $\textbf{95.6}$  & \textbf{84.8}                \\ %Max Full: 96.04, 
			\hline                                                                                                          
			Snips 			& 88.2\tnote{\ddag}	& 97.0			        & 74.9\tnote{\ddag}    & 88.2			        & \textbf{96.9}			& 74.6			                \\
			Snips+ 			& \textbf{89.3}        & \textbf{97.3}	    & \textbf{76.4}        & \textbf{88.3}        & 96.7  & \textbf{74.6}                \\
			\hline                                                                                                          
			ME	            & 82.2\tnote{\ddag}	& -					    & 63.6\tnote{\ddag}	& 81.8\tnote{\ddag}    & -					& 62.1\tnote{\ddag}			\\
			ME+	            & \textbf{82.9}	    & -					    & \textbf{64.5}		& \textbf{82.8}        & -			        & \textbf{63.3}			    \\
			\hline                                                                                                          
			MT              & 63.5\tnote{\ddag}	& -					    & 24.0\tnote{\ddag}	& 62.8\tnote{\ddag}	& -			        & 24.4\tnote{\ddag}			\\
			MT+             & \textbf{65.7}        & -	                    & \textbf{27.4}        & \textbf{65.0}        & -	                & \textbf{27.5}			    \\
			\hline                                                                                                          
			MR	            & 72.6\tnote{\dag}		& -					    & 52.8\tnote{\dag}		& 72.1\tnote{\ddag}	& -			        & 51.8\tnote{\ddag}			\\
			MR+	            & \textbf{73.0}        & -	                    & \textbf{53.4}        & \textbf{73.0}		& -			        & \textbf{52.9}			    \\
			\hline
		\end{tabular}
		\begin{tablenotes}
			\item[\dag]{\footnotesize $p<0.1$}
			\item[\ddag]{\footnotesize $p<0.01$}
		\end{tablenotes}
	\end{threeparttable}
	\caption{Mean data augmentation results on various SLU tasks tested using the slot-gated \cite{goo2018slot} SLU models. Datasets are augmented (prefixed by +) using our proposed generative model. The results have been aggregated and are tested for statistical significance. }
	\label{tab:general}
\end{table}

For other SLU models, we consider the slot-gated SLU model \cite{goo2018slot}, which incorporates the attention and the gating mechanism into the LU network. We found the model suitable for our task, as the model is reasonably complex and distinctive from our simple baseline. Furthermore, the code for running the model is publicly available and the results are readily reproducible. We were able to obtain similar or even better results on our environment (Table \ref{tab:atis-slot}). This difference might be due to differing data preprocessing methods. SLU performance is measure by (1) slot filling f1-score (evaluated using the conlleval perl script), (2) intent identification f1-sore, and (3) semantic frame formulation. f1-score measures the correctness of predicted slot labels. 

\subsection{Generative Data Augmentation Results}

In this section, we describe and present two experiments that test the GDA approach under variety of experimental settings: data scarce scenarios, varied SLU models, and varied datasets. 

\subsubsection{Data Scarce Scenario}

\begin{table}
	\centering
	\begin{threeparttable}
		\begin{tabular}{llcc}
			\hline
			Dataset 	& Model							& Slot (F1)	& Intent (F1)\\
			\hline\hline
			ATIS		& JLUVA					& 94.44     & 97.09     \\
			\hline
			%ATIS        & \begin{tabular}{@{}@{} l} Hybrid RNN \\
			%            \cite{Mesnil2015} \end{tabular} & 95.06     & -         \\
			ATIS		& BiLSTM (Baseline)& 95.34 	& 97.21      \\  
			ATIS        & Deep LSTM\tnote{a}			& 95.66 & -     \\
			ATIS		& Slot-Gated (Full)\tnote{b,d} 	& 95.66	& 96.08 \\
			ATIS        & Att. Encoder-Decoder\tnote{c} & 95.87 & \textbf{98.43} \\
			ATIS        & Att. BiRNN\tnote{c}    		& 95.98 & 98.21 \\
			\hline				
			ATIS+		& BiLSTM (Baseline)				& 95.75	& 97.54		\\
			ATIS+ 		& Slot-Gated (Full)\tnote{b,d}    & \textbf{96.04} & 96.75     \\
			\hline
		\end{tabular}
		\begin{tablenotes}
			\item[a]{\footnotesize \cite{kurata2016leveraging}}
			\item[b]{\footnotesize \cite{goo2018slot}}
			\item[c]{\footnotesize \cite{liu2016attention}}
			\item[d]{\footnotesize run on our environment}
		\end{tablenotes}
	\end{threeparttable}
	\par
	\caption{Comparisons of the best slot filling and intent detection results for the ATIS dataset.}
	\label{tab:atis-slot}
\end{table}

For the first experiment, we test whether our GDA approach performs better than the previous work 1) under the regular condition (full datasets) 2) and data scarce scenarios. We compare our model to a deterministic encoder-decoder model (Seq2Seq) proposed in \cite{kurata2016labeled}. The two decoders of the model learn to decode utterances and slot labels from an encoded representation of the utterance. %In an attempt to reproduce the results of the original models, we restrict the model from generating intents. However, we still could observe some discrepancies between our results and the results reported in the paper (Table \ref{tab:data-scarce}), possibly due to minor differences in experimental protocols. 

For the full dataset, we conduct the standard experiments with $N_G=3, N_L=3$ and $m=10000$, synthetic dataset size. For small and medium datasets, each experiment is repeated $N_L=3$ times for \textit{all} $N_T$ training splits. The final result is aggregated from $N_T \times N_L$ runs (i.e. 105 runs for ATIS-small and 27 runs for ATIS-medium). Results are presented in \ref{tab:data-scarce}.

According to the results, our approach performed better than all other baselines at the statistically significant level for small and medium datasets. The performance gain of our approach diminishes for the full dataset. This is likely due to the homogeneous nature of the ATIS dataset, leaving little room for the GDA to explore. Although we could not achieve statistically significant improvement on the full dataset, we note that our approach never experiences performance degradation for any dataset size and evaluation measure. 

\subsubsection{GDA on Other SLU Models and Datasets}

We test GDA with various combinations of SLU models and datasets (Table \ref{tab:general}). There were statistically significant improvements in language understanding performances across most datasets and SLU models. Comparing these results with the data scarcity results in Table \ref{tab:data-scarce}, we observe two trends: (1) the more difficult the dataset is to model (e.g. MIT Movie Trivia) and (2) the more expressive the SLU model, the more drastic the improvements are. For example, the improvement rate between ATIS and ATIS+ for full attention-based Slot-Gated model was only 0.39\%, whereas the improvement rate increased nearly ten-fold (3.54\%) between MIT Movie Eng and MIT Movie Eng+ for the same model. 

We also observe a positive correlation between model complexity and performance gains. For example, the performance improvement was more significant for the slot-gated model than the simple baseline model for the ATIS dataset. This suggests that the performance-boosting benefits from synthetic datasets can be more easily captured by more expressive models. This is also supported by generally better performances achieved by the slot-gated full attention model, as the full attention variant is the more complex one.

\subsection{Comparison to Other State-of-the-art Results}

In this study, we compare the best LU performance achieved by our generative approach on the ATIS task to other state-of-the-art results in literature (Table \ref{tab:atis-slot}). We chose the best performing run out of all runs carried out from the previous experiments ($N_G=3,N_L=3, m=10000$) and report its results in Table \ref{tab:atis-slot}. In the best case, our approach was able to boost the slot filling performance for the slot-gated (full) model by 0.38. Remarkably, our best results outpeformed more complex models, further supporting the idea of data-centric regularization. We also evaluate the SLU performance of JLUVA by performing deterministic inference (i.e. $\mathbf{z} = \boldsymbol{\mu}$). We find that the LU performance by itself is not competitive. This eliminates the possibility that the performance gains in our approach are attributed to JLUVA being a more expressive model and therefore acting as a teacher network.

\subsection{Ablation Studies}

In the ablation studies, we carry out two separate comparative experiments on variations of our generative model. 

\subsubsection{Sampling Methods}

The following sampling approaches are considered.

\begin{itemize}
	\item \textbf{Monte-Carlo Posterior Sampling (Ours)}: $\mathbf{z}$ is sampled from the empirical expectation of the model, which is estimated by inferring posteriors from random utterance samples. (Algorithm \ref{alg:posterior-sampling})
	\item \textbf{Standard Gaussian}: $\mathbf{z}$ is sampled from the assumed prior, the standard multivariate Gaussian.
	\item \textbf{Additive Sampling}: First, the latent representation $\mathbf{z}_\mathbf{w}$ of a random utterance $\mathbf{w}$ is sampled. Then $\mathbf{z}_\mathbf{w}$ is disturbed by a perturbation vector $\boldsymbol{\alpha} \sim \mathcal{U}\left(-0.2,0.2\right)$. It was proposed for the deterministic model in \cite{kurata2016labeled}.
\end{itemize}

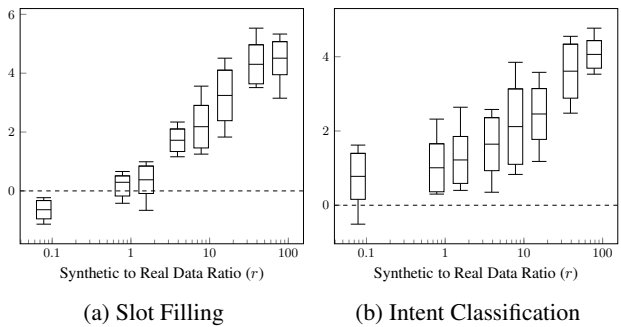
\begin{figure}
	\centering
	\begin{subfigure}[b]{0.23\textwidth}
		\resizebox{\textwidth}{!}{
			\begin{tikzpicture}
			\begin{axis} [xmode=log, log ticks with fixed point, xtick pos=left, ytick pos=left, xlabel={Synthetic to Real Data Ratio ($r$)}, ylabel={}, ylabel shift=5pt, label style={font=\large}]
			\addplot [box plot median] table {aaai2019-kmyoo-data-augmentation-ablation-slot.dat};
			\addplot [box plot box] table {aaai2019-kmyoo-data-augmentation-ablation-slot.dat};
			\addplot [box plot top whisker] table {aaai2019-kmyoo-data-augmentation-ablation-slot.dat};
			\addplot [box plot bottom whisker] table {aaai2019-kmyoo-data-augmentation-ablation-slot.dat};
			\draw[dashed] ({rel axis cs:0,0}|-{axis cs:0,0}) -- ({rel axis cs:1,0}|-{axis cs:0,0});
			\end{axis}
			\end{tikzpicture}
		}
		\caption{Slot Filling}
		\label{fig:ablation-slot}
	\end{subfigure}
	\begin{subfigure}[b]{0.23\textwidth}
		\resizebox{\textwidth}{!}{
			\begin{tikzpicture}
			\begin{axis} [xmode=log, log ticks with fixed point, xtick pos=left, ytick pos=left, xlabel={Synthetic to Real Data Ratio ($r$)}, ylabel={}, ylabel shift=5pt, label style={font=\large}]
			\addplot [box plot median] table {aaai2019-kmyoo-data-augmentation-ablation-intent.dat};
			\addplot [box plot box] table {aaai2019-kmyoo-data-augmentation-ablation-intent.dat};
			\addplot [box plot top whisker] table {aaai2019-kmyoo-data-augmentation-ablation-intent.dat};
			\addplot [box plot bottom whisker] table {aaai2019-kmyoo-data-augmentation-ablation-intent.dat};
			\draw[dashed] ({rel axis cs:0,0}|-{axis cs:0,0}) -- ({rel axis cs:1,0}|-{axis cs:0,0});
			\end{axis}
			\end{tikzpicture}
		}
		\caption{Intent Classification}
		\label{fig:ablation-intent}
	\end{subfigure}
	\par
	\caption{The impact of synthetic data to real data ratio on the relative improvements in SLU performance. The vertical axis shows the relative performance gains, compared to the non-augmented baseline (dashed horizontal lines). For each box plot, the height of the box depicts the variance and outer whiskers mark the minimum and the maximum.}
	\label{fig:ablation}
\end{figure}

The results in Table \ref{tab:data-scarce} confirm that exploratory Monte-Carlo sampling based on scaled posterior distribution ($\lambda_s=0.18$) provides the greatest benefit to the language understanding models for the ATIS and the data-scarce datasets. We note that the additive perturbation, despite its simplicity in nature, performs reasonably well compared to our approach. This suggests the exploratory sampling approaches are not only limited to Gaussian distributions. On the other hand, over-simplified and biased approximation of the prior such as standard multivariate Gaussian, could rather cause performance degradation. This also highlights the fact that the choice of sampling approach has a significant impact on the generative quality and thereby the resulting performances. 

\subsubsection{Synthetic Data Ratio}

%Would excessive augmentation degrade the SLU performance? 
To gain further insights into generative DA, we conduct regressional experiments to expose the underlying relationship between the relative synthetic data size and the performance improvements.

Let $m$ be the size of the synthetic dataset used to augment the original dataset of size $n$. The \textit{synthetic to real data ratio} $r$ is $m/n$. For each run, we conduct the standard experiment procedure ($N_G=10, N_L=5$) on a ATIS-small dataset with JLUVA as the generative model and the simple BiLSTM as the SLU model. We repeat the experiments for all $r \in \left\{0.08, 0.78, 1.56, 3.90, 7.81, 15.6, 39.06, 78.13 \right\}$. From the box plots of our results (Figure \ref{fig:ablation}), we make two observations. 
First, the maximum marginal improvement is achieved around $10 \leq r \leq 20$ for all evaluation measures. Also, the improvements appear to plateau around $r=50$. Second, The variance starts off relatively small when $r < 1$, but it quickly grows as $r$ increases and peaks around $5 \leq r \leq 20$. The variance appears to shrink again after $r > 20$. A plausible explanation for the apparent trend of the variance is that increasing $r$ enhances the chance to generate performance-boosting key utterances, until no novel instances of such utterances are samplable from the generator, at which point further increasing $r$ only increases the chance to generate already known utterances, thereby reducing the variance. This also explains the plateauing phenomenon.

% \section{Qualitative Analysis}

% %TODO: qualitative analysis introduction

% \subsection{Case Study}

% %TODO: cherry pick results
% %TODO: analyze specific cases

% - unit gaussian sampling yields homogeneous results

% \subsection{Generation Analysis}

% %TODO: more quantitative side of qualitative analysis

%\subsection{Remarks}
%Suppose that all utterance samples $\mathbf{w} \in \mathcal{D}_w$ were sampled from a true but unknown spoken language distribution $p \left( \mathbf{w} \right)$. However, due to bias introduced during erroneous data collection process or due to undersampling variance, the actual language distribution represented by the dataset the  Data augmentation is a process whereby a sufficiently realistic data model $\hat{p} \left( \mathbf{w} \right) \approx p \left( \mathbf{w} \right)$ exists such that $m$ new samples $\hat{\mathbf{w}}_1,\ldots,\hat{\mathbf{w}}_M$ drawn from $\hat{p} \left( \mathbf{w}\right)$ combined with the original dataset $\mathcal{D}

\section{Conclusion}

In this paper, we formulated the generic framework for generative data augmentation (GDA) and derived analytically the most effective sampling approach for generating performance-boosting instances from our proposed generative model, Joint Language Understanding Variational Autoencoder (JLUVA).
Based on the positive experimental results, we believe that our approach could bring immediate benefits to SLU researchers and the industry by reducing the cost of building new SLU datasets and improve performances of existing SLU models. Although our work has primarily been motivated by the data issues in SLU datasets, we would like to invite researchers to explore the potential of applying GDA in other NLP tasks, such as neural machine translation and natural language inference. Similar to the work done by \citeauthor{dao2018kernel} on the analysis of class-preserving transformative DAs using the kernel theory \cite{dao2018kernel}, our work also calls for deeper theoretical analysis on the mechanism of data-centric regularization techniques. We wish to address these issues in our future work.

%\nocite{*}

\newpage

\fontsize{9pt}{10pt} \selectfont
\bibliography{aaai2019-kmyoo-data-augmentation-short}
\bibliographystyle{aaai}

\end{document}